\documentclass{llncs}

\usepackage[a4paper,includeheadfoot,top=1.65in,bottom=1.65in,left=1.73in,hcentering]{geometry}
\usepackage[pdftex]{graphicx}
\usepackage[sectionbib]{natbib}

\usepackage[utf8]{inputenc} 
\usepackage{t1enc}
\usepackage{float}
\restylefloat{figure}
\usepackage[english]{babel}
\usepackage{amsmath}
\usepackage{graphicx}
\usepackage{booktabs}
\usepackage{tabularx}
\usepackage{xcolor}
\usepackage{hyperref}
\usepackage{multirow}

\newif{\ifhidecomments}
\hidecommentstrue

\ifhidecomments
    \newcommand{\judit}[1]{}
    \newcommand{\andras}[1]{}
    \newcommand{\dani}[1]{}
    \newcommand{\david}[1]{}
\else
    \newcommand{\judit}[1]{\textcolor{red}{[#1 ({\bf Judit})]}}
    \newcommand{\andras}[1]{\textcolor{blue}{[#1 ({\bf Andr\'as})]}} 
    \newcommand{\dani}[1]{\textcolor{orange}{[#1 ({\bf Dani})]}} 
    \newcommand{\david}[1]{\textcolor{brown}{[#1 ({\bf D\'avid})]}} 
\fi
\newcommand{\fhref}[2]{\href{#1}{#2}\footnote{\url{#1}}}
\newif{\ifhidetodo}
\hidetodotrue

\ifhidetodo
    \newcommand{\todo}[1]{}
\else
    \newcommand{\todo}[1]{\textcolor{magenta}{[TODO: #1 ]}}
\fi

\definecolor{cb_green2}{RGB}{102,194,165}
\definecolor{cb_red}{RGB}{252,141,98}
\definecolor{cb_blue}{RGB}{141,160,203}
\definecolor{cb_purple}{RGB}{231,138,195}
\definecolor{cb_green1}{RGB}{166,216,84}
\definecolor{cb_yellow}{RGB}{255,217,47}
\definecolor{cb_salmon}{RGB}{229,196,148}
\definecolor{cb_gray}{RGB}{179,179,179}

\usepackage{tikz}
\usepackage{pgffor}
\usetikzlibrary{matrix,positioning,shapes.geometric, arrows, fit}
\usetikzlibrary{calc}
\tikzstyle{module_color} = [fill=cb_gray!70]
\tikzstyle{module} = [
    rectangle, rounded corners, text centered,
    module_color,
    minimum height=.8cm, draw=black
]
\tikzstyle{large_module} = [
    module,
    text width=7cm,
    minimum width=7cm,
]

\tikzstyle{bert_module} = [
    module,
    text width=9cm,
    minimum width=10cm,
]

\tikzstyle{small_module} = [
    module,
    text width=1.5cm,
    minimum width=1.5cm,
]

\tikzstyle{sequence} = [
    minimum height=0.6cm,
    rectangle, text centered,
    draw=black, fill=cb_yellow!80,
]
\tikzstyle{target_sequence} = [
    sequence, fill=cb_red!90,
]
\tikzstyle{output} = [
    rectangle, text centered,
    draw=black, fill=cb_green2!80
]

\tikzstyle{arrow} = [thick,->,>=stealth]
\tikzset{
    *|/.style={
        to path={
            (
            perpendicular cs: 
            horizontal line through={(\tikztostart)},
            vertical line through={(\tikztotarget)}
            )
            -- (\tikztotarget) \tikztonodes
        }
    }
}

\tikzset{
    box/.style={rectangle,draw=black,thick, minimum size=0.01cm},
}

\def\rowdist{1.5}
\def\matrixheight{0.8}
\def\matrixwidth{0.4}

\newcommand*{\drawbertoutput}[2]{%
    \begin{scope}[shift={(#1-\matrixwidth/2, #2-\matrixheight/2)}]
        \foreach \x in {0,0.1,...,0.4}{
            \foreach \y in {0,0.1,...,0.8}{
                \draw [line width=0.05pt, fill=cb_yellow!75, draw=black] (\x, \y) rectangle (\x+0.1, \y+0.1);
            }%
        }%
    \end{scope}
}

\newcommand*{\drawberttargetoutput}[2]{%
    \begin{scope}[shift={(#1-\matrixwidth/2, #2-\matrixheight/2)}]
        \foreach \x in {0,0.1,...,0.4}{
            \foreach \y in {0,0.1,...,0.8}{
                \draw [fill=cb_red!90, line width=0.05pt, draw=black] (\x, \y) rectangle (\x+0.1, \y+0.1);
            }%
        }%
    \end{scope}
}

\begin{document}

\title{Evaluating Contextualized Language Models for Hungarian}
\author{Judit \'Acs\inst{1,2}, D\'aniel L\'evai\inst{3}, D\'avid M\'ark Nemeskey\inst{2}, Andr\'as Kornai\inst{2}}
\institute{
    $^1$ Department of Automation and Applied Informatics\\
    Budapest University of Technology and Economics\break
    $^2$ Institute for Computer Science and Control\\
    $^3$ Alfréd Rényi Institute of Mathematics\\
}

\maketitle

\begin{abstract}
    We present an extended comparison of contextualized language models for
    Hungarian. We compare huBERT, a Hungarian model against 4 multilingual
    models including the multilingual BERT model. We evaluate these models
    through three tasks, morphological probing, POS tagging and NER. We find
    that huBERT works better than the other models, often by a large margin,
    particularly near the global optimum (typically at the middle
    layers). We also find that huBERT tends to generate fewer subwords for
    one word and that using the last subword for token-level tasks is
    generally a better choice than using the first one.  
\end{abstract}

\section{Introduction}

Contextualized language models such BERT \citep{Devlin:2018a} drastically
improved the state of the art for a multitude of natural language processing
applications. \cite{Devlin:2018a} originally released 4 English and
2 multilingual pretrained versions of BERT (mBERT for short) that support over
100 languages including Hungarian. BERT was quickly followed by other large
pretrained Transformer \citep{Vaswani:2017} based models such as RoBERTa
\citep{Liu:2019a} and multilingual models with Hungarian support such as
XLM-RoBERTa \citep{Conneau:2019a}. Huggingface released the Transformers
library \citep{Wolf:2020}, a PyTorch
implementation of Transformer-based language models along with a repository for
pretrained models from community contribution
\footnote{\url{https://huggingface.co/models}}. This list now contains over
1000 entries, many of which are domain- or language-specific models.

Despite the wealth of multilingual and language-specific models, most
evaluation methods are limited to English, especially for the early models.
\cite{Devlin:2018a} showed that the original mBERT outperformed existing
models on the XNLI dataset \citep{Conneau:2018b}. mBERT was further evaluated
by \cite{Wu:2019} for 5 tasks in 39 languages, which they later expanded to
over 50 languages for part-of-speech tagging, named entity recognition and
dependency parsing \citep{Wu:2020a}. 

\cite{Nemeskey:2020} released the first BERT model for Hungarian named
\emph{huBERT} trained on Webcorpus 2.0 \citep[ch. 4]{Nemeskey:2020}. It uses
the same architecture as BERT base with 12 Transformer layers with 12 heads and
768 hidden dimension each with a total of 110M parameters. huBERT has a
WordPiece vocabulary with ~30k subwords.

In this paper we focus on evaluation for the Hungarian language. We compare
huBERT against multilingual models using three tasks: morphological probing, POS
tagging and NER. We show that huBERT outperforms all multilingual models,
particularly in the lower layers, and often by a large margin. We also show that
subword tokens generated by huBERT's tokenizer are closer to Hungarian
morphemes than the ones generated by the other models.

\section{Approach}

We evaluate the models through three tasks: morphological probing, POS tagging and
NER.  Hungarian has a rich inflectional morphology and largely free word order.
Morphology plays a key role in parsing Hungarian sentences.

We picked two token-level tasks, POS tagging and NER for assessing the sentence
level behavior of the models.
POS tagging is a common subtask of downstream NLP applications such as
dependency parsing, named entity recognition and building knowledge graphs.
Named entity recognition is indispensable for various high level semantic
applications.

\subsection{Morphological probing}

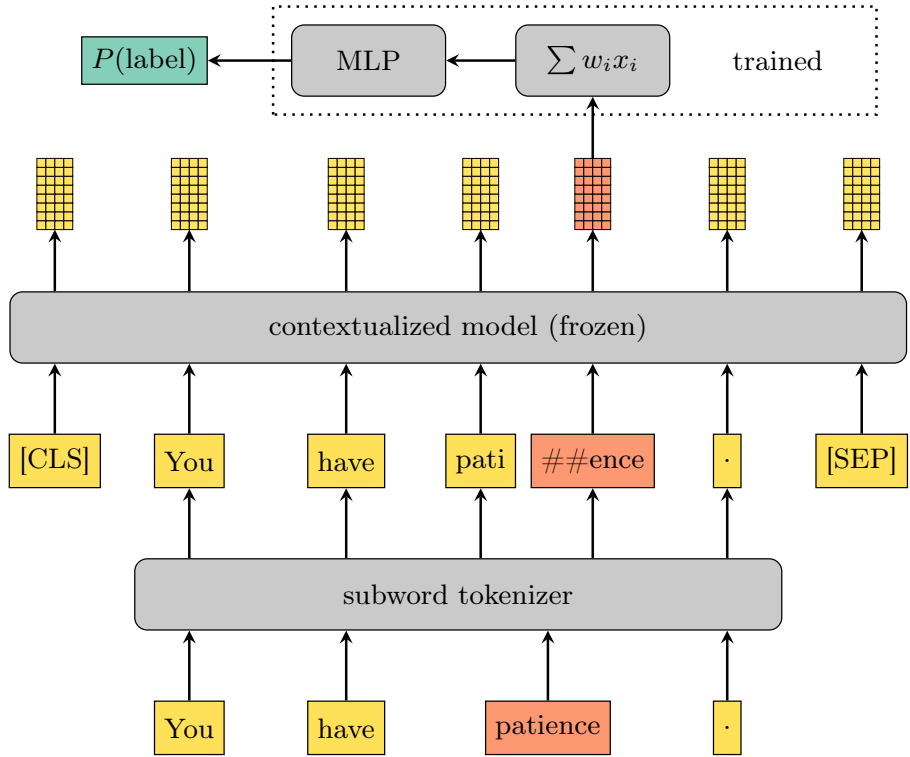
\begin{figure}[t]
    \resizebox{0.99\textwidth}{!}{
    \begin{tikzpicture}
        \node (tokenizer) at (0, 0) [large_module] {subword tokenizer};
        \node (t1) at (-3, -\rowdist) [sequence] {You};
        \node (t2) at (-1.25, -\rowdist) [sequence] {have};
        \node (t3) at (1, -\rowdist) [target_sequence] {patience};
        \node (t4) at (3, -\rowdist) [sequence] {.};
        \foreach \x in {1,...,4}
            \draw [arrow,<-] (tokenizer.south) to[*|] (t\x.north);
        \node (wp1) at (-4.5, \rowdist) [sequence] {[CLS]};
        \node (wp2) at (-3, \rowdist) [sequence] {You};
        \node (wp3) at (-1.25, \rowdist) [sequence] {have};
        \node (wp4) at (0.25, \rowdist) [sequence] {pati};
        \node (wp5) at (1.5, \rowdist) [target_sequence] {\#\#ence};
        \node (wp6) at (3, \rowdist) [sequence] {.};
        \node (wp7) at (4.5, \rowdist) [sequence] {[SEP]};
        \foreach \x in {2,...,6}
            \draw [arrow] (tokenizer.north) to[*|] (wp\x.south);
        \node[fit=(wp4)(wp5)](wp_full){};

        \node (bert) at (0, 2*\rowdist) [bert_module] {contextualized model
        (frozen)};
        \foreach \x in {1,...,7}
        \draw [arrow,<-] (bert.south) to[*|] (wp\x.north);

        \foreach \x in {-4.5,-3,-1.25,0.25,3,4.5} \drawbertoutput{\x}{3*\rowdist} ;
        \drawberttargetoutput{1.5}{3*\rowdist} ;
        \foreach \x in {-4.5,-3,-1.25,0.25,1.5,3,4.5}
            \draw [arrow] (bert.north) to[*|] (\x, 3*\rowdist-\matrixheight/2);

        \node (weight) at (1.5, 4*\rowdist) [small_module] {$\sum w_i x_i$};
        \node (mlp) at (-1, 4*\rowdist) [small_module] {MLP};
        \draw [arrow,<-] (weight.south) -- (1.5, 3*\rowdist+\matrixheight/2);
        \draw [arrow] (weight.west) -- (mlp.east);
        \node (output) at (-3.5, 4*\rowdist) [output] {$P(\text{label})$};
        \draw [arrow] (mlp.west) -- (output.east);

        \draw[dotted,thick] ($(mlp.south west)+(-0.2, -0.2)$) rectangle
        ($(weight.north east)+(2.3, 0.2)$);
        \node[] at ($(weight.east)+(1.2, 0)$) {trained};

    \end{tikzpicture}
    }

    \caption{Probing architecture. Input is tokenized into subwords and a
      weighted average of the mBERT layers taken on the last subword of
      the target word is used for classification by an MLP. Only the MLP
      parameters and the layer weights $w_i$ are trained.}
\label{fig:tikz_architecture}
\end{figure}

Probing is a popular evaluation method for black box models. Our approach is
illustrated in Figure~\ref{fig:tikz_architecture}.  The input of a probing
classifier is a sentence and a target position (a token in the sentence).  We
feed the sentence to the contextualized model and extract the representation
corresponding to the target token. We use either a single Transformer layer of
the model or the weighted average of all layers with learned weights.  We
train a small classifier on top of this representation that predicts a
morphological tag. We expose the classifier to a limited amount of training
data (2000 training and 200 validation instances).  If the classifier performs
well on unseen data, we conclude that the representation includes said
morphological information.  We generate the data from the automatically tagged
Webcorpus 2.0.  The target words have no overlap between train, validation and
test, and we limit class imbalance to 3-to-1 which resulted in filtering some
rare values. The list of tasks we were able to generate is summarized in
Table~\ref{table:morph_tasks}.

\begin{table}
	\centering
	\begin{tabular}{llcl}
        \toprule
        Morph tag & POS & \#classes & Values \\
        \midrule
        Case & noun & 18 & Abl, Acc, \dots, Ter, Tra  \\
        Degree & adj & 3 & Cmp, Pos, Sup \\
        Mood & verb & 4 & Cnd, Imp, Ind, Pot\\
        Number psor & noun & 2 & Sing, Plur \\
        Number & adj & 2 & Sing, Plur \\
        Number & noun & 2 & Sing, Plur \\
        Number & verb & 2 & Sing, Plur \\
        Person psor  & noun & 3 & 1, 2, 3\\
        Person & verb & 3 & 1, 2, 3 \\
        Tense & verb & 2 & Pres, Past \\
        VerbForm & verb & 2 & Inf, Fin \\
        \bottomrule
	\end{tabular}
    \caption{\label{table:morph_tasks} List of morphological probing tasks.}
\end{table}

\subsection{Sequence tagging tasks}

Our setup for the two sequence tagging tasks is similar to that of the
morphological probes except we train a shared classifier on top of all token
representations. Since multiple subwords may correspond to a single token (see
Section~\ref{sec:segmentation} for more details), we need to aggregate them in
some manner: we pick either the first one or the last one.\footnote{We also
experimented with other pooling methods such as elementwise max and sum but
they did not make a significant difference.}

We use two datasets for POS tagging. One is the Szeged Universal Dependencies
Treebank \citep{Farkas:2012,Nivre:2018s} consisting of 910 train, 441 validation, and
449 test sentences. Our second dataset is a subsample of Webcorpus 2 tagged
with emtsv \citep{Indig:2019} with 10,000 train, 2000 validation, and 2000 test sentences.

Our architecture for NER is identical to the POS tagging setup. We train it on
the Szeged NER corpus consisting of 8172 train, 503 validation, and 900 test
sentences.

\subsection{Training details}

We train all classifiers with identical hyperparameters. The classifiers have
one hidden layer with 50 neurons and ReLU activation. The input and the
output layers are determined by the choice of language model and the number of
target labels. This results in 40k to 60k trained parameters, far fewer than the
number of parameters in any of the language models.

All models are trained using the Adam optimizer \citep{Kingma:2014} with
$lr=0.001, \beta_1 = 0.9, \beta_2 = 0.999$. We use 0.2 dropout for
regularization and early stopping based on the development set.

\section{The models evaluated}
\label{sec:evaluation}

We evaluate 5 models.

\begin{description}
    \item[huBERT] the Hungarian BERT, is a BERT-base model with 12 Transformer
      layers, 12 attention heads, each with 768 hidden dimensions and a total
      of 110 million parameters. It was trained on Webcorpus 2.0
      \citep{Nemeskey:2020}, 9-billion-token corpus compiled from the
      Hungarian subset of \fhref{https://commoncrawl.org/}{Common Crawl}.  Its
      string identifier in Huggingface Transformers is\linebreak
      \verb|SZTAKI-HLT/hubert-base-cc|.
    \item[mBERT] the cased version of the multilingual BERT. It is a BERT-base
        model with identical architecture to huBERT.
        It was trained on the Wikipedias of the 104 largest Wikipedia
        languages.  Its string id is \linebreak \verb|bert-base-multilingual-cased|.
    \item[XLM-RoBERTa] the multilingual version of RoBERTa. Architecturally,
        it is identical to BERT; the only difference lies in the training
        regimen. XLM-RoBERTa was trained on 2TB of Common Crawl data, and it
        supports 100 languages. Its string id is \verb|xlm-roberta-base|.
    \item[XLM-MLM-100] is a larger variant of XLM-RoBERTa with 16 instead
      of 12 layers. Its string id is
      \verb|xlm-mlm-100-1280|.
      
    \item[distilbert-base-multilingual-cased] is a \textit{distilled} version
        of mBERT. It cuts the parameter budget and inference time by roughly
        40\% while retaining 97\% of the tutor model's NLU capabilities. Its
        string id is \linebreak \verb|distilbert-base-multilingual-cased|.
\end{description}

\subsection{Subword tokenization}\label{sec:segmentation}

Subword tokenization is a key component in achieving good performance on morphologically rich languages.
Out of the 5 models we compare, huBERT, mBERT and DistilBERT use the WordPiece algorithm \citep{Schuster:2012}, XLM-RoBERTa and XLM-MLM-100 use the SentencePiece algorithm \citep{Kudo:2018b}.
The two types of tokenizers are algorithmically very similar, the differences
between the tokenizers are mainly dependent on the vocabulary size per
language.
The multilingual models consist of about 100 languages, and the vocabularies per language are (not linearly) proportional to the amount of training data available per language.
Since huBERT is trained on monolingual data, it can retain less frequent subwords in its vocabulary, while mBERT, RoBERTa and MLM-100, being multilingual models, have token information from many languages, so we anticipate that huBERT is more faithful to Hungarian morphology. 
DistilBERT uses the tokenizer of mBERT, thus it is not included in this subsection.

\begin{table}
	\centering
	\begin{tabular}{l  r  r  r  r  r}
        \toprule
				    	 &huBERT & mBERT & RoBERTa & MLM-100 & \hspace{5mm} emtsv \\
        \midrule
		Languages		    & 1    & 104  & 100  & 100  & 1\\
		Vocabulary size 	& 32k  & 120k & 250k & 200k & --\\
		Entropy of first WP & 8.99 & 6.64 & 6.33 & 7.56 & 8.26\\
		Entropy of last WP  & 6.82 & 6.38 & 5.60 & 6.89 & 5.14\\
		More than one WP    &94.9\%&96.9\%&96.5\%&97.0\%&95.8\%\\ 

		Length in WP &2.8$\pm$1.4& 3.9$\pm$1.8& 3.2$\pm$1.4 & 3.5$\pm$1.5 & 3.1$\pm$1.1\\

		Length of first WP&4.3$\pm$3.0&2.7$\pm$1.9& 3.5$\pm$2.7 & 3.1$\pm$2.0 & 5.2$\pm$2.4\\

		Length of last WP&3.8$\pm$2.9& 2.6$\pm$1.8 & 3.1$\pm$2.2 & 2.8$\pm$1.8 & 2.7$\pm$1.7\\
		
		Accuracy to emtsv&0.16 & 0.05 & 0.14 & 0.08 & 1.00 \\
		Accuracy to emtsv in first WP& 0.41 & 0.26 & 0.44 & 0.33 & 1.00 \\
		Accuracy to emtsv in last WP& 0.43 & 0.41 & 0.47 & 0.39 & 1.00 \\

        \bottomrule
	\end{tabular}
	\caption{Measures on the train data of the POS tasks. 
			The length of first and last WP is calculated in
                        characters, while the word length is calculated in
                        WPs. DistilBERT data is identical to mBERT.}\label{table:subword}
\end{table}

As shown in Table~\ref{table:subword}, there is a gap between the Hungarian and multilingual models in almost every measure.
mBERT's shared vocabulary consists only of 120k subwords for all 100 languages while huBERT's vocabulary contains 32k items and is uniquely for Hungarian.
Given the very limited inventory of mBERT, only the most frequent Hungarian words are represented as a single token, while longer Hungarian words are segmented, often very poorly.
The average number of subwords a word is tokenized into is 2.77 in the case of huBERT, while all the other models have significantly higher mean length.
This does not pose a problem in itself, since the tokenizers work with a given dictionary 
size and frequent words need not to be segmented into subwords. 
But in case of words with rarer subwords, the limits of smaller monolingual vocabulary can be observed, as shown in the following example: \textit{szállítójárművekkel} `with transport vehicles'; \textit{szállító-jármű-vek-kel} `transport-vehicle-\textsc{pl}-\textsc{ins}' for huBERT, \textit{sz-ál-lí-tó-já-rm-ű-vek-kel} for mBERT, which found the affixes correctly (since affixes are high in frequency), but have not found the root `transport vehicle'.

\begin{figure}
	\centering
	\includegraphics[width=0.6\textwidth]{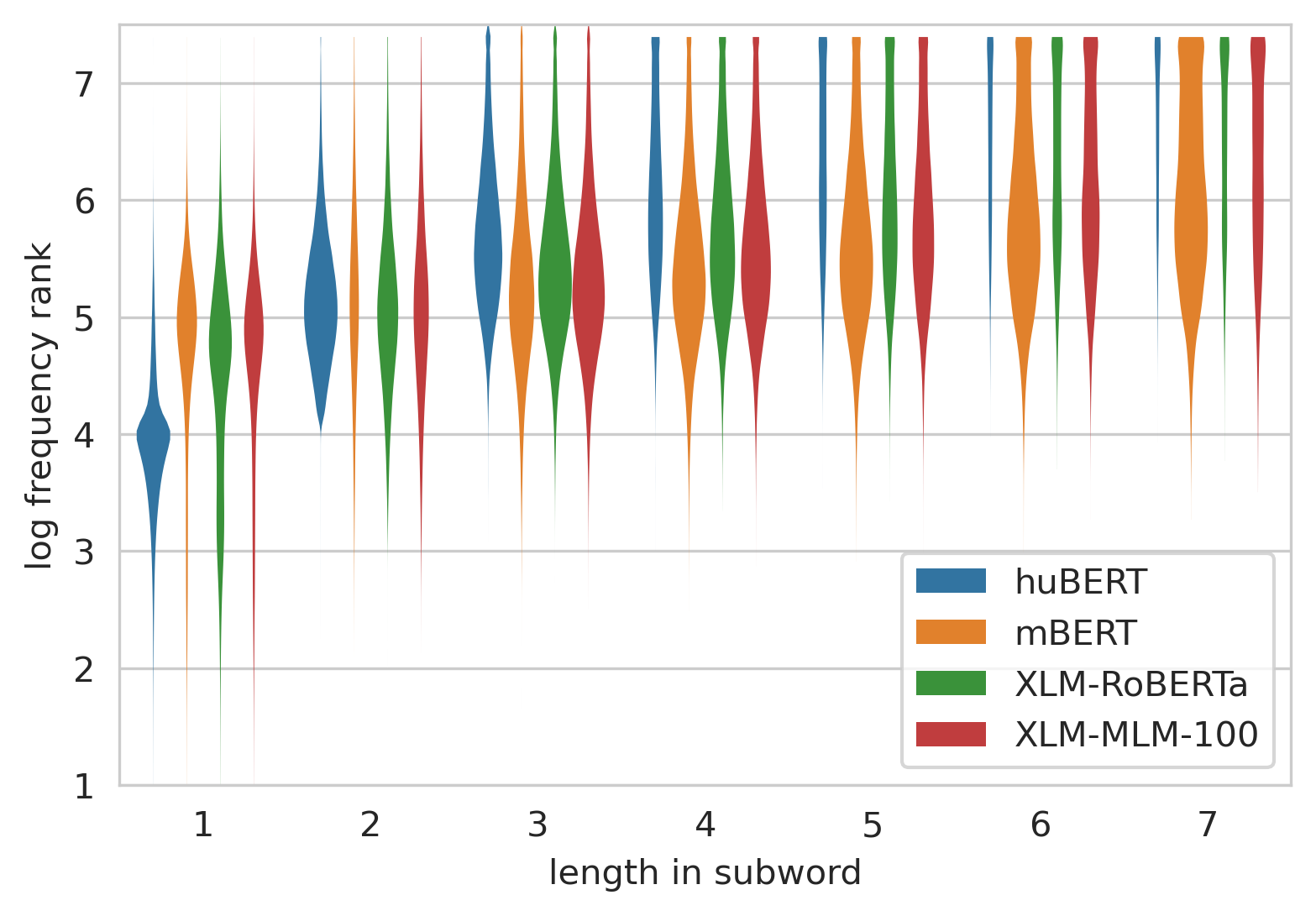}
	
	\caption{Distribution of length in subword vs. log frequency rank. 
		The count of words for one subword length is proportional to the size of the respective violin. }
	\label{fig:subword}
\end{figure}

Distributionally, huBERT shows a stronger Zipfian distribution than any other model, as shown in Figure~\ref{fig:subword}. 
Frequency and subword length are in a linear relationship for the huBERT model, while in case of the other models, the subword lengths does not seem to be correlated the log frequency rank. 
The area of the violins also show that words typically consist of more than 3 subwords for the multilingual models, contrary to the huBERT, which segments the words typically into one or two subwords.

\section{Results}

We find that huBERT outperforms all models in all tasks, often with a large
margin, particularly in the lower layers. As for the choice of subword pooling
(first or last) and the choice of layer, we note some trends in the following
subsections.

\subsection{Morphology}

\begin{figure}
    \makebox[\textwidth][c]{\includegraphics[clip,trim={0 0 0 0},width=1.2\textwidth]{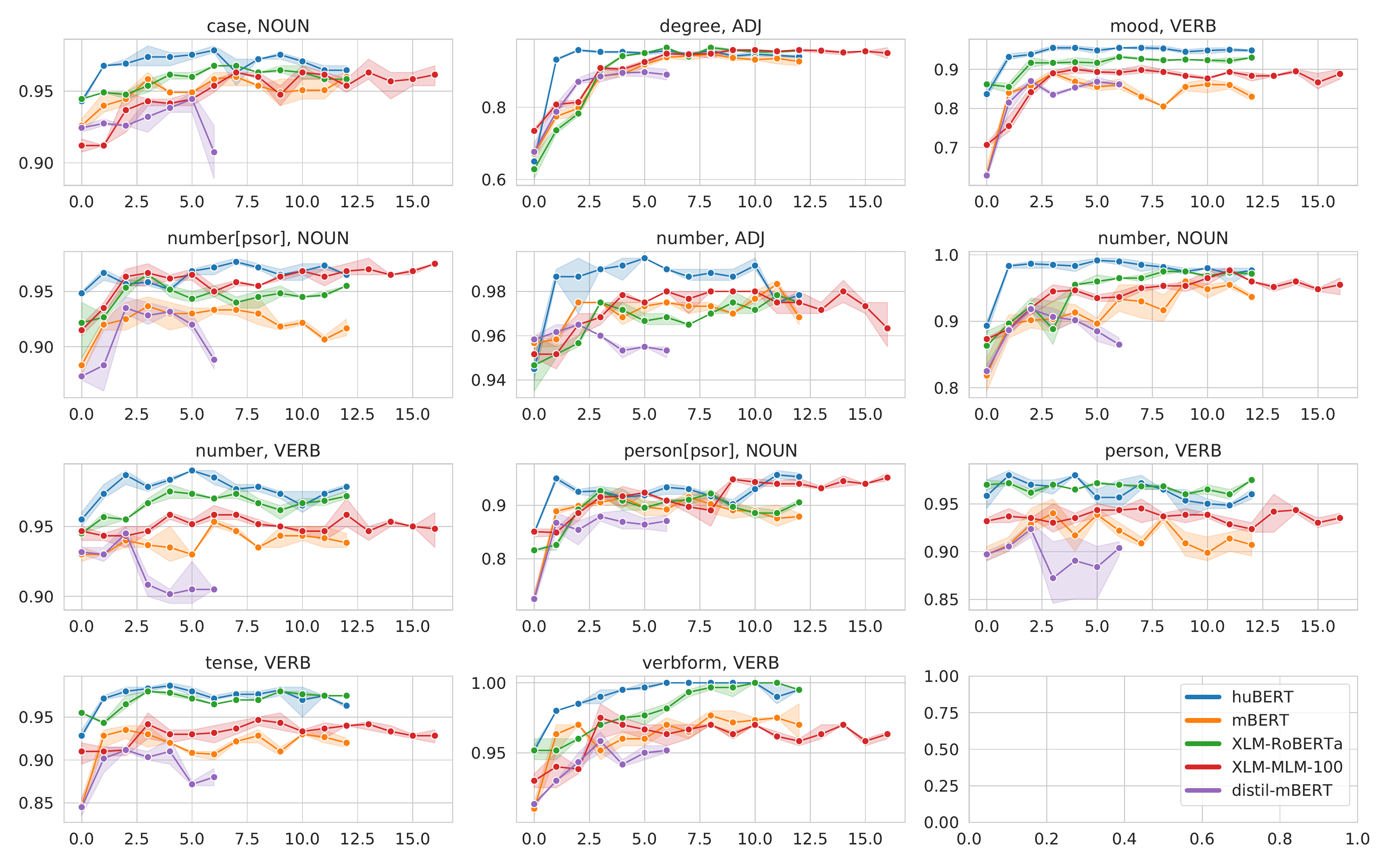}}
    \caption{\label{fig:line_morph_layerwise} The layerwise accuracy of
    morphological probes using the last subword. Shaded areas represent
confidence intervals over 3 runs.} 
\end{figure}

The last subword is always better than the first subword except for a few cases
for degree ADJ.  This is not surprising because superlative is marked with a
circumfix and it is differentiated from comparative by a prefix.  The rest of
the results in this subsection all use the last subword.

huBERT is better than all models, especially in the lower layers in
morphological tasks, as shown in Figure~\ref{fig:line_morph_layerwise}.
However, this tendency starts at the second layer and the first layer does not
usually outperform the other models.  In some morphological tasks huBERT
systematically outperforms the other models: these are mostly the simpler noun and
adjective-based probes.  In possessor tasks (tagged \verb|[psor]| in
Figure~\ref{fig:line_morph_layerwise}) XLM models are comparable to huBERT, while
mBERT and distil-mBERT generally perform worse then huBERT.  In verb tasks
XLM-RoBERTa achieves similar accuracy to huBERT in the higher layers,
while in the lower layers, huBERT tends to have a higher accuracy.

\begin{figure}
    \makebox[\textwidth][c]{\includegraphics[clip,trim={0 0 0 0},width=1.2\textwidth]{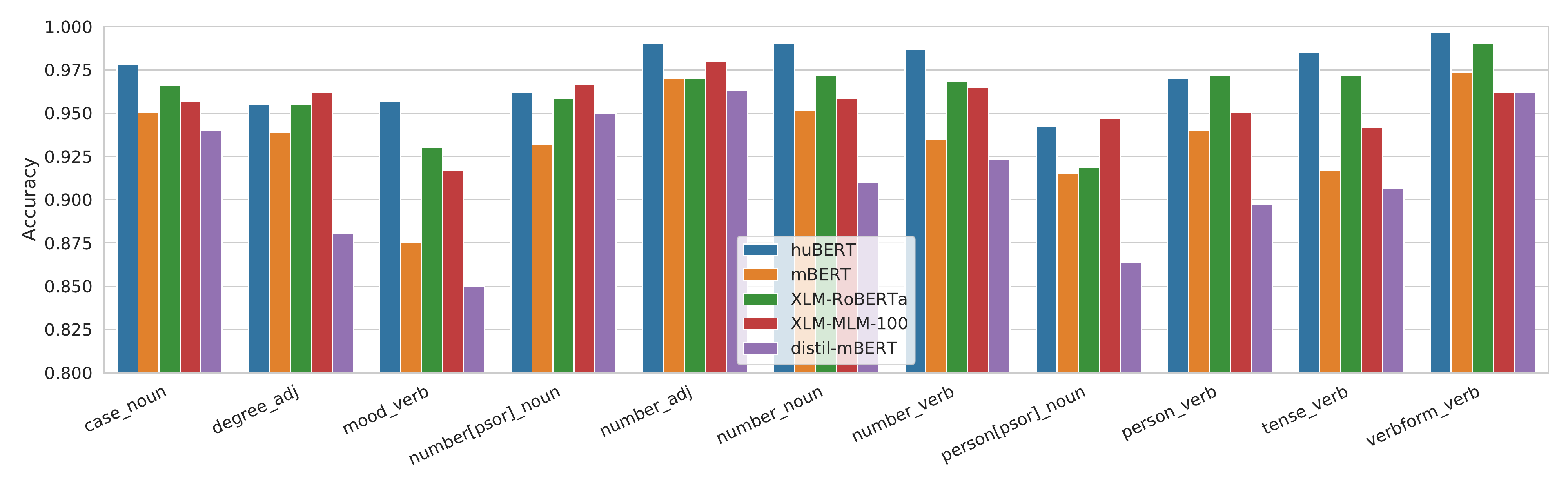}}
    \caption{\label{fig:bar_morph_summary} Probing accuracy using the weighted
        sum of all layers. }
\end{figure}

HuBERT is also better than all models in all tasks when we use the weighted
average of all layers as illustrated by Figure~\ref{fig:bar_morph_summary}. The
only exceptions are adjective degrees and the possessor tasks. A possible
explanation for the surprising effectiveness of XLM-MLM-100 is its higher layer
count.

\subsection{POS tagging}

\begin{figure}
    \includegraphics[clip,trim={0 0 0 0},width=.99\textwidth]{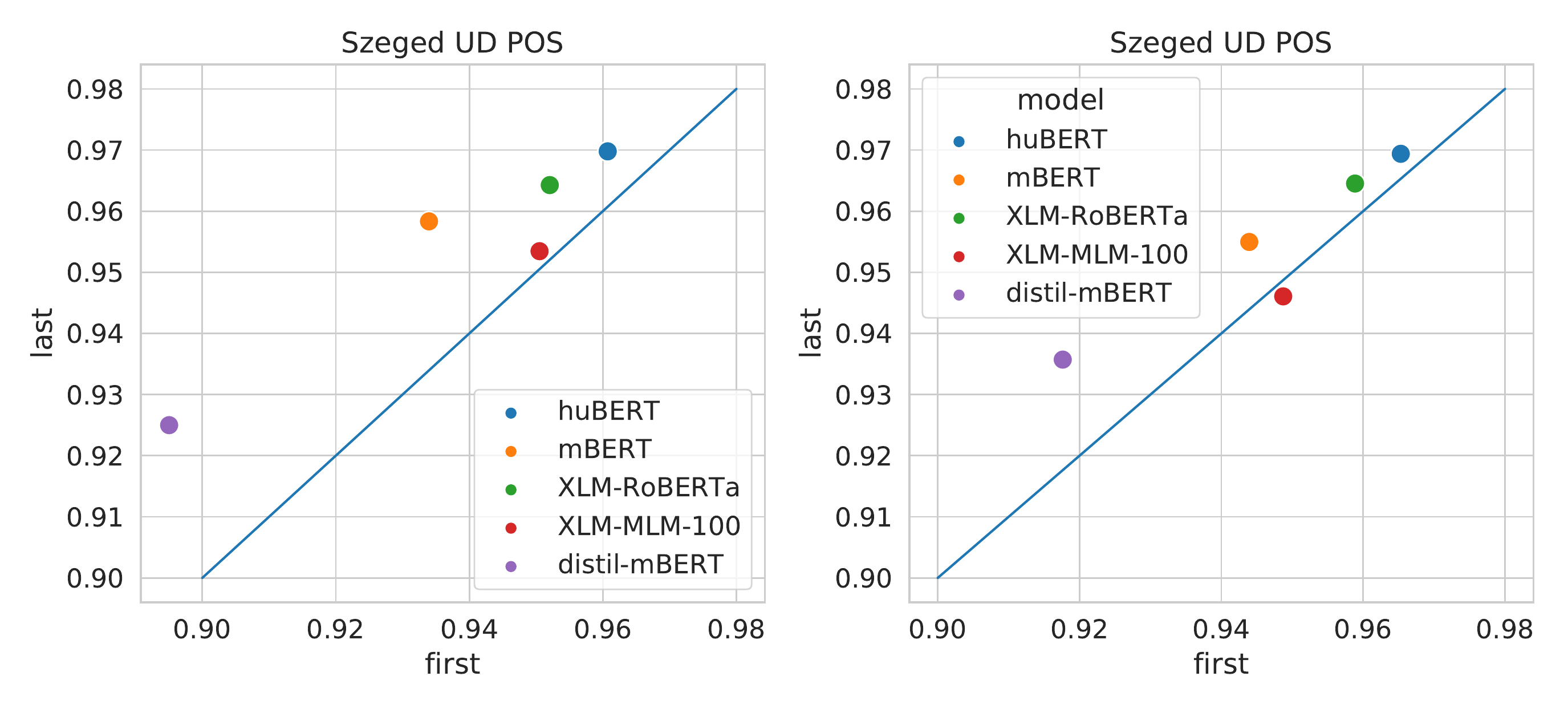}
    \caption{\label{fig:scatter_pos} POS tag accuracy on Szeged UD and on the Webcorpus 2.0 sample}
\end{figure}

Figure~\ref{fig:scatter_pos} shows the accuracy of different models on the
gold-standard Szeged UD and on the silver-standard data created with emtsv.

Last subword pooling always performs better than first subword pooling.  As in
the morphology tasks, the XLM models perform only a bit worse than huBERT.
mBERT is very close in performance to huBERT, unlike in the morphological
tasks, while distil-mBERT performs the worst, possibly due to its far lower
parameter count.

\begin{figure}
    \includegraphics[clip,trim={0 0 0 0},width=.99\textwidth]{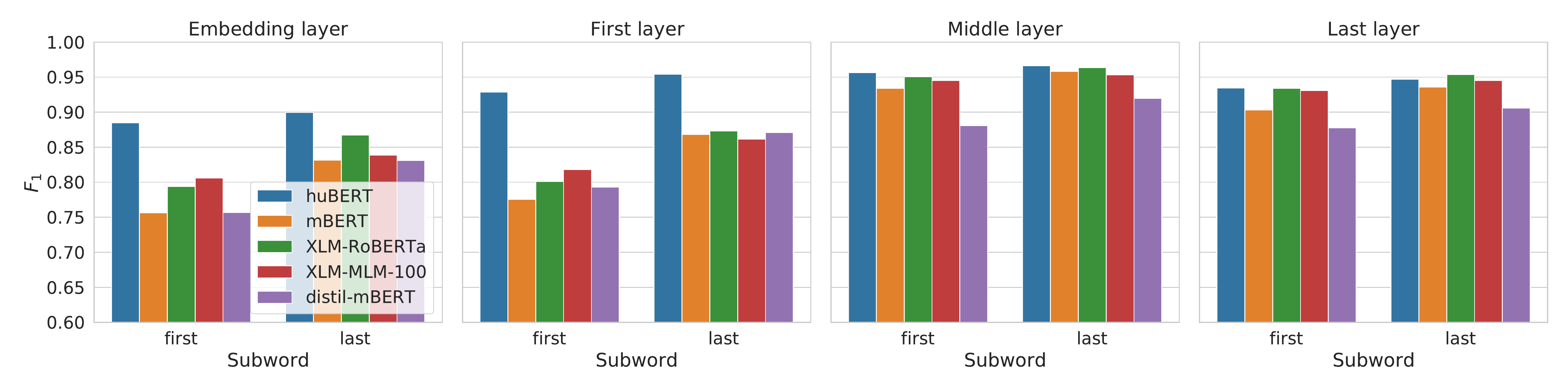}
    \caption{\label{fig:bar_szeged_pos_by_layer} Szeged POS at 4 layers:
    embedding layer, first Transformer layer, middle layer, and highest layer.}
\end{figure}

We next examine the behavior of the layers by relative position.\footnote{We
only do this on the smaller Szeged dataset due to resource limitations.} The
embedding layer is a static mapping of subwords to an embedding space
with a simple positional encoding added. Contextual information is not
available until the first layer. The highest layer is generally used as the
input for downstream tasks. We also plot the performance of the middle layer. As
Figure~\ref{fig:bar_szeged_pos_by_layer} shows, the embedding layer is the
worst for each model and, somewhat surprisingly, adding one contextual layer only
leads to a small improvement. The middle layer is actually better than the
highest layer which confirms the findings of \cite{Tenney:2019b} that BERT
rediscovers the NLP pipeline along its layers, where POS tagging is a mid-level
task. As for the choice of subword, the last one is generally better, but the
gap shrinks as we go higher in layers.

\begin{figure}
    \includegraphics[clip,trim={0 0 0 0},width=.99\textwidth]{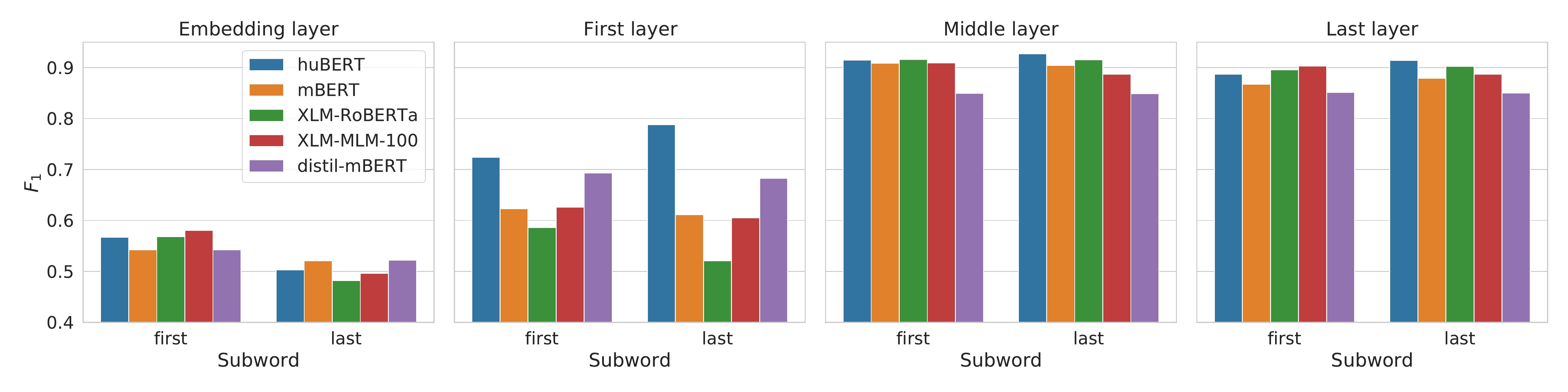}
    \caption{\label{fig:bar_ner_low_middle_high} NER $F_1$ score at the lowest,
    middle and highest layers.}
\end{figure}

\subsection{Named entity recognition}

In the NER task (Figure~\ref{fig:bar_ner_low_middle_high}), all of the
models perform very similarly in the higher layers, except for distil-mBERT
which has nearly 3 times the error of the best model, huBERT. The closer we
get to the global optimum, the clearer huBERT's superiority becomes. Far away
from the optimum, when we use only the embedding layer, first subword is
better than last, but the closer we get to the optimum (middle and last layer),
the clearer the superiority of the last subword choice becomes.

\section{Related work}

Probing is a popular method for exploring blackbox models. \cite{Shi:2016} was
perhaps the first one to apply probing classifiers to probe the syntactic
knowledge of neural machine translation models. \cite{Belinkov:2017} probed NMT
models for morphology. This work was followed by a large number of similar
probing papers \citep{Belinkov:2017,Adi:2017,Hewitt:2019,Liu:2019,Tenney:2019,Warstadt:2019,Conneau:2018,Hupkes:2018}.
Despite the popularity of probing classifiers, they have theoretical
limitations as knowledge extractors \citep{Voita:2020}, and low quality of
silver data can also limit applicability of important probing techniques
such as canonical correlation analysis \citep{Singh:2019},

Multilingual BERT has been applied to a variety of multilingual tasks such as
dependency parsing \citep{Kondratyuk:2019} or constituency
parsing\citet{Kitaev:2019}. mBERT's multilingual capabilities have been
explored for NER, POS and dependency parsing in dozens of language by
\citet{Wu:2019} and \citet{Wu:2020a}. The surprisingly effective multilinguality
of mBERT was further explored by \citet{Dufter:2020}.

\section{Conclusion}

We presented a comparison of contextualized language models for Hungarian. We
evaluated huBERT against 4 multilingual models across three tasks,
morphological probing, POS tagging and NER. We found that huBERT is almost
always better at all tasks, especially in the layers where the optima are
reached. We also found that the subword tokenizer of huBERT matches Hungarian
morphological segmentation much more faithfully than those of the multilingual
models.  We also show that the choice of subword also matters. The last
subword is much better for all three kinds of tasks, except for cases where
discontinuous morphology is involved, as in circumfixes and infixed plural
possessives \citep{Antal:1963,Melcuk:1972}. Our data, code and the full result
tables are available at \url{https://github.com/juditacs/hubert_eval}.

\section*{Acknowledgements}
This work was partially supported by National Research, Development and
Innovation Office (NKFIH) grant \#120145: ``\textit{Deep Learning of
  Morphological Structure}'', by National Excellence Programme
2018-1.2.1-NKP-00008: ``\textit{Exploring the Mathematical Foundations of
  Artificial Intelligence}'', and by the Ministry of Innovation and the
National Research, Development and Innovation Office within the framework of
the Artificial Intelligence National Laboratory Programme. L\'evai was
supported by the NRDI Forefront Research Excellence Program KKP\_20 Nr. 133921
and the Hungarian National Excellence Grant 2018-1.2.1-NKP-00008.

\renewcommand\bibname{References}
\bibliographystyle{splncsnat_en}
\bibliography{ml}

\end{document}